\documentclass[sigconf]{acmart}
\usepackage{graphicx} 
\usepackage{hyperref} 

\usepackage{amssymb}
\usepackage{float}
\usepackage{placeins}
\usepackage{subcaption} 
\usepackage{forest}
\usepackage{graphicx}
\usepackage{booktabs}
\usepackage{multirow}
\usepackage{pifont}
\usepackage{array}
\usepackage{caption}
\usepackage{rotating}  
\usepackage{color}
\usepackage{xcolor}
\usepackage[table]{xcolor}
\usepackage[dvipsnames]{xcolor} 
\usepackage{makecell}

\usepackage{algorithm}
\usepackage{algpseudocode}

\usepackage{enumitem}

\usepackage[most]{tcolorbox}

\usepackage{pifont}
\renewcommand\footnotetextcopyrightpermission[1]{}

\AtBeginDocument{%
  }

\setcopyright{acmlicensed}
\copyrightyear{2026}
\acmYear{2026}
\acmDOI{XXXXXXX.XXXXXXX}

\acmConference[Conference]{}{2026}

\begin{document}

\title[GroupGPT: An Agentic Framework for Multi-User Chat Assistant]{GroupGPT: A Token-efficient and Privacy-preserving Agentic Framework for Multi-User Chat Assistant}

\author{Zhuokang Shen}
\authornote{Both authors contributed equally to this research.}
\affiliation{%
  \institution{East China Normal University}
  \city{Shanghai}
  \country{China}
}

\author{Yifan Wang}
\authornotemark[1]
\affiliation{%
  \institution{East China Normal University}
  \city{Shanghai}
  \country{China}
}

\author{Hanyu Chen}
\affiliation{%
  \institution{University of Nottingham Ningbo}
  \city{Ningbo}
  \country{China}
}

\author{Yunhang Shen}
\affiliation{%
  \institution{Xiamen University}
  \city{Xiamen}
  \country{China}
}

\author{Wenxuan Huang}
\affiliation{%
  \institution{East China Normal University}
  \city{Shanghai}
  \country{China}
}

\author{Gaoqi He}
\affiliation{%
  \institution{East China Normal University}
  \city{Shanghai}
  \country{China}
}

\author{Jiao Xie}
\affiliation{%
  \institution{East China Normal University}
  \city{Shanghai}
  \country{China}
}

\author{Rongrong Ji}
\affiliation{%
  \institution{Xiamen University}
  \city{Xiamen}
  \country{China}
}

\author{Shaohui Lin}
\authornote{Corresponding author. (Email: shlin@cs.ecnu.edu.cn).}
\affiliation{%
  \institution{East China Normal University}
  \city{Shanghai}
  \country{China}
}

\renewcommand{\shortauthors}{Shen et al.}


\begin{abstract}
Recent advances in large language models (LLMs) have enabled increasingly capable chatbots. 
However, most existing systems focus on single-user settings and do not generalize well to multi-user group chat interactions, where agents require more proactive and accurate intervention under complex, evolving contexts.
Existing approaches typically rely on LLMs for both intervention reasoning and response generation, leading to high token consumption, limited scalability, and potential privacy risks.
To address these challenges, we propose \textbf{GroupGPT}, a token-efficient and privacy-preserving agentic framework for multi-user chat assistant. 
GroupGPT adopts an edge-cloud model collaboration architecture to decouple intervention timing from response generation, enabling efficient and accurate decision-making while preserving user privacy through on-device processing of sensitive information. 
The framework also supports multimodal inputs, including memes, images, videos, and voice messages.
To support evaluation of timing accuracy and response quality, we further introduce \textbf{MUIR}, a benchmark dataset for multi-user chat assistant intervention reasoning.
MUIR contains 2,500 annotated group chat segments with intervention labels and rationales. 
We evaluate a range of models on MUIR, spanning from open-source to proprietary variants, including both LLMs and their smaller counterparts.
Extensive experiments demonstrate that GroupGPT generates accurate and well-timed responses, achieving an average score of 4.72/5.0 in LLM-based evaluation, and is well-received by users across diverse group chat scenarios.
Moreover, GroupGPT reduces the token usage by up to $3\times$ compared to baselines, while providing privacy sanitization of user messages before cloud transmission. 
Code is available at: \url{https://github.com/Eliot-Shen/GroupGPT}.
\end{abstract}

\maketitle


\begin{figure}[t]
\centering
\includegraphics[width=\columnwidth]{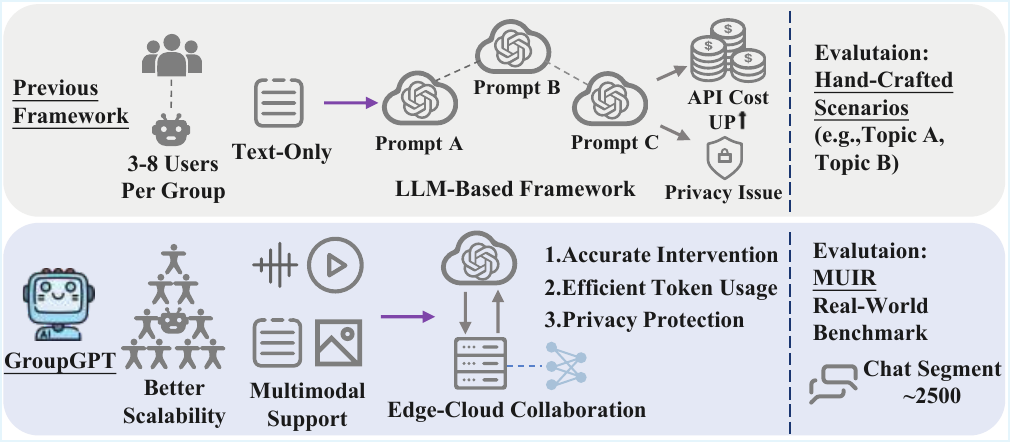}
\caption{Comparison between GroupGPT and prior frameworks. 
GroupGPT mitigates existing limitations via a multi-agent, edge-cloud design.
We further introduce MUIR, a real-world benchmark dataset with annotated rationales for more comprehensive and objective evaluation.
}
\label{fig:contrast}
\end{figure}

\section{Introduction}
The domain of chatbot research has experienced a remarkable boom in recent times, driven by rapid advances in LLMs and their increasing deployment in real-world applications such as customer service, education, and personal assistants. 
Nonetheless, current research \cite{budzianowski2019hello,hosseini2020simple,su2022multi,wang2022task,yang2021ubar} is predominantly concentrated on interactions involving a single user.
While these systems demonstrate strong capabilities in single-user interactions, their extension to multi-user group chat settings remains underexplored.
In industry settings, most existing group chatbots tend to focus on passive responses, operating in a rule-driven manner, rather than demonstrating proactive participation, such as enterprise bots in WeChat Work or Discord bots that primarily respond to predefined commands or triggers.
This scarcity of studies concerning group conversation chatbots significantly impedes their deployment in complex, multi-user settings, including collaborative activities like brainstorming sessions, debate and emotional communication.

Recently, multi-user chatbots have begun gaining momentum in both academia and industry(e.g., ChatGPT’s group feature and TikTok group chat bots). 
However, as shown in Figure~\ref{fig:contrast}, existing open-source frameworks \cite{mao2024multi,jacniacki2025humanlike,lee2025map} exhibit several key limitations:
(1) High token consumption: Modular designs rely on multiple LLM-driven components, resulting in costly inference and poor scalability in high-frequency group chats. 
(2) Insufficient privacy protection: Raw conversational data is often sent directly to cloud-based LLMs, raising data security and governance issues. 
(3) Lack of multimodal support: Most frameworks only focus on text, lacking effective integration of images, audio, and videos common in real-world group interactions.
(4) Limited evaluation methodology: reliance on manually constructed scenarios and user studies leads to subjective and insufficiently diverse assessments.

To address the above issues, we propose GroupGPT, a multi-agent based chatbot framework tailored for real-world social group. 
GroupGPT can understand complex multimodal social content, including user-shared images, memes, videos, and voice messages.  
It adopts a edge-cloud model collaborative design. 
To protect users' private information, it introduce a sensitive information detection and rewriting model in the inference stage. 
Furthermore, it incorporate an intervention judge model that filters out the majority of scenarios where system intervention is unnecessary, which significantly reduces the LLM's token footprint, thereby improving scalability in high-volume group-chat environments.

To support quantitative evaluation of intervention behavior, we propose an innovative data curation pipeline that constructs MUIR, a benchmark dataset for multi-user chat assistant intervention reasoning, which contains 2500 group-chat segments with annotated rationales.
In evaluation, the chatbot instantiated under the framework achieves an overall score of 4.72/5.0 in response quality as evaluated by an LLM-as-a-judge methodology, and is ultimately well received by the majority of participants.

Our contributions are summarized as follows:
\begin{itemize}[left=0pt, topsep=1pt]
\item We propose GroupGPT, a novel group chatbot framework composed of several specialized sub-agents, including an Intervention Judge, a Privacy Transcriber, a Multimodal Processor, and a Final Respondent.
\item To the best of our knowledge, MUIR is the first benchmark dataset for group chat intervention with human-annotated rationales. We further propose an innovative data curation pipeline for constructing this dataset.
\item Comprehensive experiments show that GroupGPT delivers high-quality responses, achieves efficient token consumption, and provides strong privacy protection.
\end{itemize}
\vspace{-0.3em}

\section{Related Work}
\subsection{Multi-user Chatbots.}
Mainstream dialogue system research has primarily centered on single-user chatbots, with extensive exploration of pre-training or fine-tuning LLMs for task-oriented dialogue systems. 
Several works \cite{budzianowski2019hello,hosseini2020simple,su2022multi,wang2022task,yang2021ubar} have employed LLMs, pre-trained or fine-tuned on dialogue data, to develop dialogue models or chatbots for various domains, such as travel ticket booking or restaurant reservation. 
In contrast, academic research on multi-party or multi-user dialogue has predominantly concentrated on foundational conversational understanding tasks using multi-party conversation datasets, including addressee recognition, speaker identification, response selection, and response generation \cite{gu2023gift,gu2021mpc,ouchi2016addressee,song2022supervised,zhang2018addressee}.

Unlike single-user chatbots, limited research on group chatbots limits their applications to tasks such as brainstorming sessions and debates.
To bridge this gap, the MUCA framework \cite{mao2024multi} was the first to formalize the ``3W'' design dimensions—``What'' to say, ``When'' to respond, and ``Who'' to answer—introducing conversational strategies for group discussions.  
Building on this foundation, the HUMA framework \cite{jacniacki2025humanlike} further enables chatbots to exhibit more natural, human-like behavior by leveraging conversational strategies and response timing.
Beyond conversational strategy design, other works have explored group chatbots from application perspectives.
For example, MAP \cite{lee2025map} proposes a multi-agent framework for multi-user personalization, through a reflection–analysis–feedback workflow to model user preferences.
Similarly, Social-RAG \cite{wang2025social} introduces a group agent for paper recommendations that retrieves social signals from group interactions to socially ground generation.
Besides, the work \cite{liu2025proactive} has explored proactive agents with “inner thoughts” mechanisms and
\cite{karahodvza2025conceptual} proposes a framework that models group structures and interaction patterns.

Parallel to academic progress, industry has also started deploying proactive group-oriented conversational agents.  
For example, OpenAI has introduced a group chat feature in the web version of ChatGPT, while ByteDance has explored group interaction agents within TikTok's chat groups, and Tencent has integrated group AI assistants into the Yuanbao app.
Recently, the viral open-source project OpenClaw~\cite{openclaw} can also be deployed into group chats of social apps such as WhatsApp, Telegram, and Feishu. 
In contrast to prior work, our approach emphasizes a unified design that simultaneously optimizes proactivity, decision intelligence, efficiency, privacy preservation, and multimodal understanding.
\vspace{-0.5em}

\subsection{Privacy issues in Conversational Agents.}
Conversational agents built on LLMs introduce two major types of privacy vulnerabilities \cite{zhang2024s}
The first relates to conventional data security and protection issues, such as unauthorized data access, personal information trafficking, and weaknesses in the interaction pipeline that could be exploited for cyberattacks, data leaks, or ransomware \cite{kshetri2023cybercrime}.
The second involves risks unique to LLMs stemming from memorization \cite{carlini2022quantifying,carlini2021extracting,nasr2023scalable,zhang2023counterfactual}.
Prior research has demonstrated that LLMs may inadvertently retain and reproduce sensitive training data, enabling low-cost extraction attacks—for example, prompting the model to repeatedly output "poem" can elicit verbatim memorized content \cite{nasr2023scalable}.
These risks are exacerbated in real-world settings, where users routinely overshare personal information during interactions \cite{mireshghallah2024trust,zhang2024s}, and where user prompts may be used for model training or fine-tuning, often without users being aware of opt-out mechanisms \cite{zhang2024s}.

To address these issues, prior research has largely adopted model-centered strategies.
During training, methods such as data sanitization \cite{kandpal2022deduplicating,lison2021anonymisation} and differentially private  \cite{li2021large,yu2021differentially} reduce exposure to sensitive data. 
Post-training approaches (e.g., knowledge unlearning \cite{jang2023knowledge}) remove information associated with particular token sequences.
At inference time, privacy risks can be further controlled through personally identifiable information(PII) detection and rewriting \cite{dou2024reducing,ngong2025protecting,zhou2025rescriber}, which we adopt in our framework.

\begin{figure*}[t]
  \includegraphics[width=\textwidth]{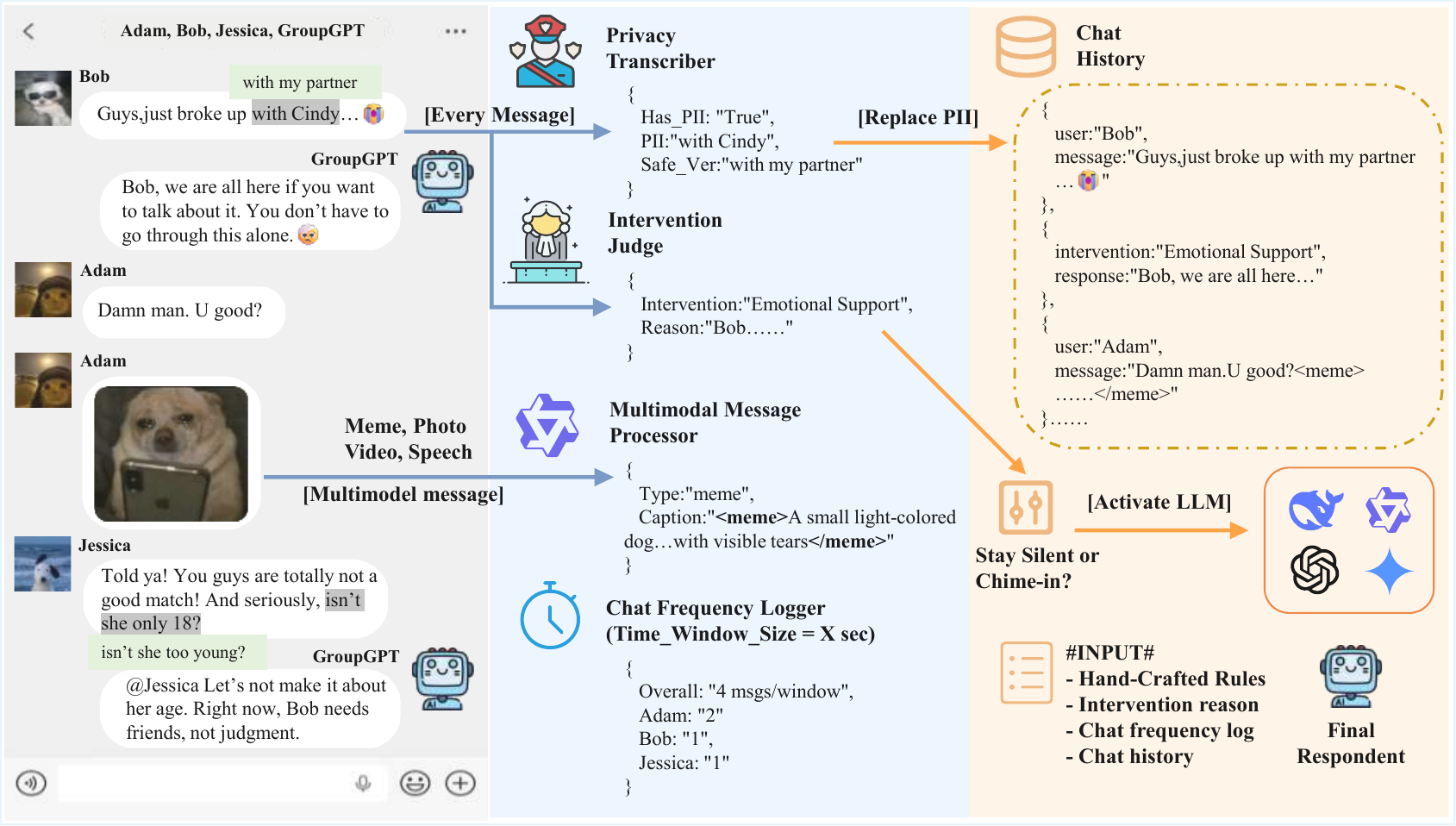}
  \caption{GroupGPT can identify the right moment to chime in, rewrite sensitive personally identifiable information(gray part), understand multimodal content such as memes, and actively participate in group discussions.}
  \label{fig:framework}
\end{figure*}

\section{Framework Architecture}
\subsection{Preliminary and Design Challenges}
\textbf{Design Challenges.}
Designing an intelligent group chatbot is a non-trivial task that presents numerous challenges.
In our work, we focus on addressing the key challenges:
\begin{itemize}[left=0pt, topsep=1pt]
\item \textbf{Accurate Intervention Timing.}
The assistant must precisely determine \emph{when} to intervene, requiring fine-grained contextual reasoning over dynamic multi-user interactions.
Premature intervention may disrupt natural dialogue flow, while delayed responses reduce utility.
The strict inference latency requirements also constrain the complexity of the overall framework design.

\item \textbf{Efficient Intervention Decision Mechanisms.}
Naive or rule-based triggering strategies can incur substantial API token overhead and delayed or missed timely interventions. 
For example, prior frameworks like MUCA \cite{mao2024multi} adopt fixed-interval evaluation (e.g., invoking an LLM every three messages to determine whether intervention is needed). 

\item \textbf{Multimodal Message Understanding.}
Existing group-chat research mainly focuses on text-only conversations. 
However, real-world social platforms involve complex multimodal interactions, including images, videos, voice messages, and stickers. 

\item \textbf{Lack of Public Group-Chat Data.}  
Due to privacy concerns and data governance restrictions, large-scale multi-user group chat logs are seldom released for public research use. 
Meanwhile, real-world group chats involve pervasive internet slang, abbreviations, and multimodal content, leading to high linguistic variability.
Such complexity is difficult for LLM-generated synthetic data to faithfully replicate, resulting in a gap between synthetic and real distributions.
Consequently, the lack of high-quality annotated real data hinders effective training and reliable benchmarking, thereby hindering systematic progress in this domain.


\item \textbf{Cloud-Based Privacy Risks.}
Deploying LLM-based assistants via cloud APIs introduces potential privacy exposure for sensitive group-chat content. 
Since group conversations may contain personal or organizational information, transmitting chat logs to external servers raises security and data governance concerns. 
\end{itemize}
These challenges jointly require a framework that balances responsiveness, efficiency, privacy, and contextual reasoning.

\textbf{Group-Chat Environment.}
We begin by formalizing the multi-user group-chat environment and the group-chat assistant task.
Let $\mathcal{C}$ denote the complete message stream of a multi-user group chat, represented as a temporally ordered sequence of utterances: $\mathcal{C} = \{u_1, u_2, \dots, u_T\}$,
where each utterance $u_i$ is associated with a speaker identity $s_i \in \{1,\dots,P\}$ and contains text or captions of images, videos or voice messages.
For the $i$-th utterance, we define the corresponding conversation context as the prefix of the message stream up to that point:
$\mathcal{C}_i = \{u_1,\dots,u_i\}$.
To support context-aware reasoning, we define a sliding window of the $N$ most recent utterances:$\mathcal{U}_{N,i} = \{u_{i-N+1}, \dots, u_i\}$.
We consider two window configurations:  
(i) a short-term window $\mathcal{U}_{N^{\mathrm{sw}}, i}$ modeling local conversational dynamics, and  
(ii) a long-term window $\mathcal{U}_{N^{\mathrm{lw}}, i}$ capturing the broader thematic context.
We denote the LLM–based agent as $p_{\theta}$ with parameters $\theta$, and introduce two lightweight auxiliary agents:  
an \emph{intervention judge} $p_{\text{ij}, \phi}$ with parameters $\phi$, and a \emph{privacy transcriber} $p_{\text{pt}, \psi}$ with parameters $\psi$. 

\textbf{Group-Chat Assistant Task.}
A Group-Chat Assistant operates over a live multi-user message stream and determines \emph{(1) what to say}, \emph{(2) when to intervene} and \emph{(3) who to respond}.
Formally, given the $i$-th utterance, the assistant must output an intervention action:
$ a \in \mathcal{A}$.
We define six types of interventions as the action set $\mathcal{A}$:
\begin{itemize}[left=0pt, topsep=1pt]
    \item \textbf{Stay Silent} — stay silent when the conversation flows naturally.
    \item \textbf{Emotional Support} — provide comfort, empathy or humor.
    \item \textbf{Offering Suggestion} — propose ideas, alternative perspectives or solutions relevant to the conversation.
    \item \textbf{Fact Correction} — gently correct factual mistakes or misinformation in the discussion.
    \item \textbf{Knowledge Enrichment} — enrich the conversation with background information, context or related facts that help others understand the topic better.
    \item \textbf{Style Balancing} — adjust tone, politeness or conversational style and resolve interpersonal conflicts to maintain group harmony.
\end{itemize}
These categories are derived from common conversational roles observed in real-world group chats. 

\subsection{The Proposed GroupGPT}
The proposed GroupGPT comprises five core components, as depicted in Fig.\ref{fig:framework}.
These components operate cooperatively during the conversation, enabling GroupGPT to minimize token consumption, protect user privacy, and maintain high-quality engagement in dynamic group chats.
The overall inference pipeline is described in Algorithm \ref{groupgpt_pipeline}, where each component is introduced as belows: 

\textbf{Intervention judge.}
The intervention judge continuously monitors the group-chat stream and determines whether an intervention is necessary at the current moment.
It is implemented as a lightweight language model $p_{\text{ij}, \phi}$ (e.g., Qwen-3-4B~\cite{yang2025qwen3}), instruction–tuned on our \textbf{MUIR} dataset.
Upon the arrival of the $i$-th utterance, the model takes as input the short-term window $\mathcal{U}_{N^{\mathrm{sw}}, i}$ and infers the appropriate intervention action $a \in \mathcal{A}$. 
This design avoids frequent invocation of LLMs, significantly reducing token consumption compared to prior interval-based triggering strategies. 

\textbf{Privacy transcriber.}
The privacy transcriber is implemented as a lightweight language model $p_{\text{pt}, \psi}$ (e.g., Llama-3.2-3B~\cite{grattafiori2024llama}), fine-tuned on existing privacy-annotated datasets to perform message-level sanitization. 
It processes the incoming message $u_i$ independently and first detects whether personally identifiable information (PII) is present. 
If PII is found, the model identifies which parts of the message contain sensitive information and provides equivalent transformations to generalize them while preserving utility. 
Formally, the transcriber rewrites the original message as $\widetilde{u}_i$, which is the privacy-preserving version of $u_i$. 
The privacy transcriber can operate in parallel with the intervention judge. 
This module ensures that sensitive user information is abstracted before being processed by large models, thereby reducing privacy leakage risks. 

\textbf{Multimodal message processor.}
The Multimodal message processor converts non-textual chat content (e.g., images, memes, videos, and voice messages) into structured textual representations.
For each message, it identifies its type and uses the corresponding model: images and memes are first classified and then captioned using designed prompts, while videos and speech are directly captioned by a specific multimodal model.
All outputs are wrapped within type-specific tags (e.g., \texttt{<meme>...</meme>}, \texttt{<image>...</image>}, \texttt{<video>...</video>}, \texttt{<audio>...</audio>}).
By unifying heterogeneous inputs into tagged text, the final response agent can operate on a consistent textual context, requiring only a text-based LLM for final response generation and thus improving efficiency. 

\begin{algorithm}[t]
\caption{GroupGPT online inference pipeline}
\label{groupgpt_pipeline}
\small
\begin{algorithmic}[1]
\State \textbf{Input:} Message stream $\mathcal{C} = \{u_1, \dots, u_T\}$, 
short-term window size $N^{sw}$, long-term window size $N^{lw}$
\State \textbf{Output:} Assistant intervention response $\{r_i\}$

\State Initialize short-term buffer $\mathcal{U}^{sw}$ (raw messages)
\State Initialize long-term buffer $\widetilde{\mathcal{U}}^{lw}$ (sanitized messages)
\State Initialize Chat Frequency Logger $\mathcal{F}$

\While{group chat is active}
    \State Receive new message $u_i$

    \State \textbf{/* Step 1: Multimodal Processing */}
    \If{$u_i$ is multimodal (image / meme / video / audio)}
        \State $u_i \gets$ MultimodalMessageProcessor$(u_i)$
    \EndIf

    \State Update short-term buffer:
    \State $\mathcal{U}^{sw} \gets \text{AppendAndSlide}(\mathcal{U}^{sw}, u_i, N^{sw})$

    \State \textbf{/* Step 2: Asynchronous Processing */}
    \State \textbf{Async Branch A (Intervention Judge):}
    \State \quad $a \gets p_{\text{ij}, \phi}(\mathcal{U}^{sw})$

    \State \textbf{Async Branch B (Privacy Transcriber):}
    \State \quad $\widetilde{u}_i \gets p_{\text{pt}, \psi}(u_i)$

    \State \textbf{/* Synchronization Barrier */}
    \State Wait until both $a$ and $\widetilde{u}_i$ are ready.

    \State Update long-term sanitized buffer:
    \State $\widetilde{\mathcal{U}}^{lw} \gets 
    \text{AppendAndSlide}(\widetilde{\mathcal{U}}^{lw}, \widetilde{u}_i, N^{lw})$

    \State \textbf{/* Step 3: Update Chat Flow Signals */}
    \State $\mathbf{z}_i \gets \mathcal{F}.\text{UpdateAndCompute}(u_i)$

    \State \textbf{/* Step 4: Final Response Generation */}
    \If{$a_i \neq \texttt{StaySilent}$}
        \State $r_i \gets p_\theta(
        \widetilde{\mathcal{U}}^{lw}, a, \mathbf{z}_i, \mathbf{p})$

        \State $u^{int}_i \gets 
        \langle \text{intervention: } a,\;
        \text{response: } r_i \rangle$

        \State Insert $u^{int}_i$ into conversation stream as $u_{T+1}$
        \State Output Intervention response $r_i$
    \EndIf

\EndWhile
\end{algorithmic}
\end{algorithm}

\textbf{Chat frequency logger.}
The chat frequency logger records conversational flow rates within a fixed time window $\Delta t$ (e.g., one minute).
For the $i$-th utterance with timestamp $t_i$, the logger computes the total number of messages and the per-user message counts that occur within the interval $[t_i - \Delta t,\, t_i]$.
These statistics provide auxiliary signals reflecting group-chat activeness and user participation patterns, which are later used by the final respondent. 

\textbf{Final respondent.}
If the intervention judge selects an intervention action $a$ other than \texttt{staysilent}, the final respondent is activated.
The final respondent is instantiated with an LLM to generate the final assistant response.
Its inputs include: 
(1) a set of hand-crafted rules that specify intervention actions under different conditions; 
(2) a task-specific prompt that governs the assistant’s behavioral constraints; 
(3) the chat frequency log of the chat group;
(4) the intervention action $a$ produced by $p_{\text{ij}, \phi}$ and
(5) the sanitized chat history output by $p_{\text{pt}, \psi}$.
The final respondent integrates these signals to produce a contextually appropriate response while maintaining safety and efficiency in group chat environments.

\section{MUIR: Intervention Reasoning Dataset for the Multi-User group chat}
\label{MUIR}
Due to privacy concerns, there is currently a lack of publicly available, high-quality multi-user group chat datasets in both academia and industry. 
Existing resources \cite{zhang2018personalizing,xu2022beyond,zang2020multiwoz,eric2020multiwoz,budzianowski2018multiwoz} fail to capture the complexity and richness of group communication, as they are typically constrained by small scale, outdated content, text-only formats and narrow topic coverage.  
Although LLMs offer a viable pathway for synthesizing conversational data, synthetic datasets often exhibit limited modality diversity and suffer from significant domain discrepancies between generated and natural interactions.

To address these limitations, we construct \textbf{MUIR}, a high-quality benchmark dataset specifically designed for studying intervention reasoning in multi-user group chats.
MUIR is built upon real-world human group chat logs, collected with appropriate anonymization. 
The dataset is annotated through a collaborative pipeline combining LLM labeling and human verification to ensure annotation reliability.
In total, MUIR contains 2,500 group chat segments, randomly split into 2,000 training instances and 500 test instances. 
After training on MUIR, our lightweight intervention judge model achieves performance that surpasses existing advanced large language models on the group-chat intervention task.

\textbf{Data collection.}
We recruited 30 volunteers from diverse backgrounds and instructed them to use several open-source chat log extraction tools to collect group chat data from their social applications, and all collected conversations were conducted in English.
Before data collection, all group members were informed that the data would be used solely for research purposes and would undergo strict anonymization and content filtering.
Data collection was conducted only after obtaining explicit consent from all participants within each group.
In total, we collected chat logs from nearly 50 group chats, covering a wide range of topics and interaction scenarios.
These include, but are not limited to, daily life sharing, technology discussions, fandom communities, art and creativity, pets, sports, programming, academic, emotional support, wellness and healing, cooking, and peer assistance. 

\textbf{Dataset construction process.}
Based on the collected long-term chat logs from nearly 50 group chats, we construct the dataset following the procedure outlined in Algorithm \ref{ij_dataset}.
In the preprocessing stage, we employ the multimodal message processor to convert all non-textual content into text captions.
Next, we segment the continuous chat streams using a sliding window strategy as depicted in Algorithm \ref{alg:gen_intervention}. 
Specifically, we define a long-term context window of size $W$ and introduce an overlap $O$ (set to $\left\lfloor \frac{W}{5} \right\rfloor$) between consecutive segments. 
The introduction of overlap mitigates context fragmentation at segment boundaries and enables the model to better capture cross-turn dependencies.
Each segmented chat window is then fed into GPT-4o~\cite{hurst2024gpt} with carefully designed prompts. 
The model is instructed to identify intervention types (label), generate rationales and responses, while also recognizing the exact positions of interventions using message-level identifiers (IDs).

\begin{algorithm}[t]
\caption{Intervention judge training data construction}
\label{ij_dataset}
\small
\begin{algorithmic}[1]

\State \textbf{Input:} 
Long chat history $\mathcal{C} = \{u_1,\dots,u_T\}$,
long-term window size $W$,
overlap size $O$,
short-term window size $S$,
decision range $X$

\State \textbf{Output:}
Training dataset $\mathcal{D}$

\State $\mathcal{C}^{text} \gets$ \Call{MultimodalToText}{$\mathcal{C}$}

\State $\mathcal{I} \gets$ \Call{GenerateAnnotations}{$\mathcal{C}^{text}, W, O$}

\State $\mathcal{D} \gets$ \Call{ConstructTrainingSet}{$\mathcal{C}^{text}, \mathcal{I}, S, X$}

\State \Return $\mathcal{D}$

\end{algorithmic}
\end{algorithm}

\begin{algorithm}[t]
\caption{Generate annotations}
\label{alg:gen_intervention}
\small
\begin{algorithmic}[1]

\Function{GenerateAnnotations}{$\mathcal{C}^{text}, W, O$}

\State $\mathcal{I} \gets \emptyset$
\State step $\gets W - O$
\State $i \gets 1$

\While{$i \le T$}

    \State $i_{end} \gets \min(i+W-1, T)$
    \State window $\gets \{u_i,\dots,u_{i_{end}}\}$

    \State window$_{json} \gets$ \Call{AddIDAndFormatJSON}{window}

    \State $\mathcal{I} \gets$ LLM(window$_{json}$, prompt) \\
    \Comment{Each intervention contains: $\langle$position(id), label, reason, response$\rangle$}

    \State $i \gets i + step$

\EndWhile

\State \Return $\mathcal{I}$

\EndFunction

\end{algorithmic}
\end{algorithm}

\begin{algorithm}[h]
\caption{Construct training set}
\label{alg:construct_ij}
\small
\begin{algorithmic}[1]

\Function{ConstructTrainingSet}{$\mathcal{C}^{text}, \mathcal{I}, S, X$}

\State $\mathcal{D} \gets \emptyset$
\State $i \gets 1$

\While{$i + S - 1 \le T$}

    \State window$_{raw} \gets \{u_i,\dots,u_{i+S-1}\}$

    \State $\mathcal{I}_{window} \gets$ interventions within $[i, i+S-1]$

    \State window $\gets$ \Call{InsertInterventionTags}{window$_{raw}$, $\mathcal{I}_{window}$}
    \hspace{4em}\Comment{Each tag contains only $\langle$label, reason$\rangle$}

    \State $R \gets [i+S-X,\; i+S-1]$ \Comment{R denotes decision range.}

    \State $\mathcal{I}_R \gets$ interventions inside $R$

    \If{$\mathcal{I}_R = \emptyset$}
        \State label $\gets$ StaySilent
        \State next index $\gets i+S$
    \Else
        \State select closest intervention to $(i+S-1)$
        \State label $\gets$ selected intervention type
        \State next index $\gets$ index after selected intervention
    \EndIf

    \State sample $\gets \langle$ window, label $\rangle$

    \State $\mathcal{D} \gets \mathcal{D} \cup \{$sample$\}$

    \State $i \gets$ next index

\EndWhile

\State \Return $\mathcal{D}$

\EndFunction

\end{algorithmic}
\end{algorithm}

\begin{table*}[h]
\centering
\caption{Performance comparison of different models on the MUIR benchmark.
The best result in each metric is bold and the second is underlined.}  
\begin{tabular}{c|c|c|cccc|c}
\toprule
& & & \multicolumn{5}{c}{\textbf{MUIR Dataset}} \\
\cmidrule{4-8}
\textbf{Method} & \textbf{Model} & \textbf{Size} & \multicolumn{2}{c}{Chime-in Reason} & \multicolumn{2}{c|}{Chime-in Timing} &  \multirow{2}{*}{Weighted}\\
\cmidrule{4-5} \cmidrule{6-7}
& & & Acc & Macro-F1 & Acc & F1 & \\
\midrule
\rowcolor{gray!15}
\multirow{2}{*}{Baseline}
& Random Guess & N/A & \textcolor{gray!100}{0.1721} & \textcolor{gray!100}{0.1552} & \textcolor{gray!100}{0.5926} & \textcolor{gray!100}{0.7280} & \textcolor{gray!100}{0.4120} \\
\rowcolor{gray!15} 
\multirow{-2}{*}{Baseline} & Human Evaluator & N/A & \textcolor{gray!100}{0.8859} & \textcolor{gray!100}{0.8687} & \textcolor{gray!100}{0.8642} & \textcolor{gray!100}{0.8913} 
& \textcolor{gray!100}{0.8775}\\
\midrule
\multirow{4}{*}{\shortstack{LLM + Prompt}} 
& DeepSeek-V3.2~\cite{liu2025deepseek} & 685B & 0.8122 & 0.7280 & 0.6728 & 0.7125 & 0.7314 \\  
& Qwen3-Max~\cite{yang2025qwen3} & N/A & 0.8333 & 0.7498 & 0.6358 & 0.6704 & 0.7223 \\
& Gemini-2.5-Pro~\cite{comanici2025gemini} & N/A & 0.8327 & 0.7466 & 0.7366 & 0.7929 & 0.7772 \\
& GPT-4o~\cite{hurst2024gpt} & N/A & \textbf{0.8716} & \textbf{0.8494} & 0.5802 & 0.5920 & 0.7233 \\
\midrule
\multirow{3}{*}{\shortstack{Embedding Model \\ + KNN}} 
& Gte-large-en-v1.5~\cite{zhang2024mgte} & 0.4B & 0.3560 & 0.3261 & 0.7428 & 0.8065 & 0.5579\\
& Bge-m3~\cite{chen2024m3} & 0.5B & 0.2860 & 0.2447 & 0.6646 & 0.7352 & 0.4826 \\
& Jina-embedding-v3~\cite{sturua2024jinaembeddingsv3multilingualembeddingstask} & 0.6B & 0.3045 & 0.2531 & 0.7160 & 0.7703 & 0.5110\\
\midrule
\multirow{8}{*}{SLM + Fine-Tuning} 
& Gemma-2-it~\cite{team2024gemma} & 2B & 0.7941 & 0.7027 & 0.7768 & 0.8395 & 0.7783\\
& Qwen-2.5-Instruct~\cite{qwen2.5} & 3B & \underline{0.8628} & \underline{0.8102} & 0.7536 & 0.8232 & \textbf{0.8125} \\
& Llama-3.2-Instruct~\cite{grattafiori2024llama} & 3B & 0.8095 & 0.7283 & 0.7780 & 0.8460 & 0.7905 \\
& Phi-4-Mini-Instruct~\cite{abouelenin2025phi} & 3.8B & 0.7985 & 0.6870 & 0.7407 & 0.8125 & 0.7597 \\
& Qwen-3~\cite{yang2025qwen3} & 4B & 0.7859 & 0.7182 & \textbf{0.8340} & \textbf{0.8867} & 0.8062 \\
& Qwen-2.5-Instruct~\cite{qwen2.5} & 7B & 0.8069 & 0.6217 & 0.7324 & 0.8181 & 0.7448 \\
& Llama-3.1-Instruct~\cite{grattafiori2024llama} & 8B & 0.8284 & 0.7732 & \underline{0.7847} & \underline{0.8535} & \underline{0.8100}  \\
& Qwen-3~\cite{yang2025qwen3} & 8B & 0.8134 & 0.6759 & 0.5380 & 0.5503 & 0.6444 \\
\bottomrule
\end{tabular}
\label{tab:response_quality}
\end{table*}

\textbf{Training segment construction.}
Given the annotated long-form group chat logs, a key challenge lies in how to segment them into training instances suitable for learning the intervention policy.
Specifically, the expert model is designed to determine whether an intervention is needed after the latest message in a given context window.  
However, constructing such training samples introduces issues related to label leakage and inference latency.

As shown in Algorithm \ref{alg:construct_ij}, we first insert annotated interventions into each window of length $S$ (the short-term window size), where only the \textit{label} and \textit{reason} are retained.
The reason provides an interpretable rationale from the LLM that helps the model learn intervention decisions. 
The response is excluded, as it is generated by the final respondent at inference time; requiring the intervention model to wait for a complete response would introduce unnecessary latency.
We then introduce a decision range $X$ to determine the final supervision label. 
By setting $X$ to $5$, we explicitly control the proportion of \texttt{stay silent} instances, making it the second most frequent label.
Finally, the above process is repeated using a sliding window strategy to construct the full training dataset.
As a result, each training sample consists of $S$ historical messages as input, which may include past intervention labels and their corresponding rationales, while the output is the intervention label and rationale at the current decision point.
This formulation enables effective utilization of rich group chat logs while avoiding label leakage across training samples.

\noindent\textbf{Test set Construction.}
The test set is split from the training corpus. 
We further employ three human annotators to perform a second round of annotation and correction on each test sample. 
For certain cases, we retain two valid labels, as these intervention types may naturally co-occur—for instance, providing emotional support while offering suggestions.
The overall annotation strategy intentionally favors over-intervention to reduce the risk of missed interventions. 
The final intervention decision can then be flexibly controlled at the final respondent stage.

\section{Experiment}
In this section, we first evaluate the intervention performance of various models on the MUIR test set. 
We then build a group chat system supported by our chatbot and conduct user study to assess GroupGPT's performance across different topics and group sizes. 
Furthermore, we measure the token consumption of GroupGPT, perform ablation studies, and adopt the LLM-as-a-judge methodology \cite{liu2023g} to evaluate the quality of chatbot's responses.
\subsection{Experimental Setups}
\textbf{GroupGPT Implementation.}
We adopt Qwen-3-4B as $p_{\text{ij}, \phi}$ and Llama-3.2-Instruct-3B as $p_{\text{pt}, \psi}$.
We train $p_{\text{pt}, \psi}$ using the dataset from \cite{dou2024reducing}.
For multimodal annotation, we employ Qwen-2.5-32B to caption images and videos (sampled at 1FPS). 
Audio data is transcribed using Qwen3-ASR-Flash.
The final response is produced by GPT-4o.
For training, the short-term window size $N^{sw}$ is set to 20, while the long-term window size $N^{lw}$ is set to 50.
All models are trained using LoRA\cite{hu2022lora} on 2 A6000 GPUs (48GB), with a batch size of 16. 
The learning rate is set to 2e-4 with a warmup ratio of 0.1.
We fix a global random seed of 42 to ensure the reproducibility. \\
\textbf{Evaluation Metrics.}
To evaluate the accuracy of model interventions, we adopt Acc, Macro-F1, and F1 score as metrics. 
The weighted score is computed as the arithmetic mean of reasoning and timing performance.
In the user study, we assess users' overall evaluation of GroupGPT through a questionnaire survey.
For token usage efficiency, we quantify cumulative encoding token length.

\subsection{MUIR Evaluation}
Table \ref{tab:response_quality} compares the performance of various models on the MUIR benchmark. 
The \textbf{Chime-in Reason} task evaluates a model’s ability to accurately identify the underlying reason for intervention when it decides to chime in. 
The \textbf{Chime-in Timing} task, formulated as a binary classification problem, measures whether the model can correctly recognize situations where it should remain silent.

We introduce two baselines for comparison. 
The random guess serves to reflect the inherent difficulty and label distribution of the dataset. 
In addition, we recruited three human evaluators to conduct a consensus-based assessment on the test set. 
Their results consistently outperform all automated models across evaluation metrics, demonstrating the high quality and reliability of the MUIR dataset, and highlighting the challenge posed by these tasks. 

For existing powerful LLMs, after in-context learning and prompt engineering, they perform well on the Chime-in Reason task. 
Specifically, GPT-4o’s accuracy and Macro-F1 scores are only 1.4\% and 1.9\% lower than the human level, respectively. 
However, we observe that their performance on the Chime-in Timing task is relatively low, with accuracy not exceeding 67\%. 
This is likely that such large models generally exhibit conservative behavior and fail to achieve the level of over-intervention required by our task.

Moreover, we experimented with text embedding models with fewer than 1B parameters. 
Using the MUIR training set, we applied a KNN-based classification on the test set. 
Results show that while these lightweight embedding models achieve moderate performance on the Chime-in Timing task, their performance on the Chime-in Reason task is poor, with accuracy generally around 30\%.

Subsequently, we evaluate small language models (SLMs) fine-tuned on the MUIR training set. 
These models significantly outperform large proprietary models such as GPT-4o on the Chime-in Timing task, effectively achieving the goal of proactive intervention decision-making, and are therefore well-suited to serve as an Intervention Judge. 
Notably, Qwen-2.5-Instruct-3B achieves 86.3\% accuracy and 81.0\% Macro-F1 on the Chime-in Reason task, and attains the best overall score among all evaluated models.
Meanwhile, Qwen-3-4B achieves the best performance on the Chime-in Timing task, reaching 83.4\% accuracy and 88.7\% F1.

\subsection{User Study}
\textbf{Study design.}
We recruited 30 participants primarily from universities and online platforms for our user study.
Eligibility criteria included being fluent in English, regularly using social apps or websites, and being at least 18 years old.
We first designed six primary discussion topics, including sports, academic studies, daily communication and sharing, gaming, emotional and mental well-being, and debate on a given proposition (whether the development of AI is beneficial or harmful to humanity). 
Participants were assigned into groups of five based on their topic preferences, and each group was required to accumulate at least 300 messages in their discussions. 
We introduced an LLM-only baseline, controlled solely via prompts, which simultaneously performed both intervention judgment and response generation. 
To facilitate comparison with baseline, we swapped the discussion topics between two groups (academic studies and debate), conducting additional discussions totaling over 500 messages. 
Its token consumption and inference latency were measured.
Finally, all 30 participants were brought together into a single channel for a free-form discussion, with a requirement of exceeding 1,500 total messages.
Each group chat was assigned a chatbot to participate in the conversation.
All group discussions were completed within a span of several days.

\begin{table}[t]
\centering
\caption{GPT-4 evaluation of 300 GroupGPT response samples (1--5 scale).}
\resizebox{\columnwidth}{!}{
\begin{tabular}{lcccccc}
\toprule
\textbf{Dimension} & \textbf{Avg} & \textbf{1} & \textbf{2} & \textbf{3} & \textbf{4} & \textbf{5} \\
\midrule
Relevance   & 4.74 & 2 (0.7\%)  & 8 (2.7\%)  & 14 (4.7\%) & 29 (9.7\%)  & 247 (82.3\%) \\
Coherence   & 4.79 & 1 (0.3\%)  & 6 (2.0\%)  & 11 (3.7\%) & 25 (8.3\%)  & 257 (85.6\%) \\
Fluency     & 4.90 & 0 (0.0\%)  & 2 (0.7\%)  & 5 (1.7\%)  & 13 (4.3\%)  & 280 (93.3\%) \\
Helpfulness & 4.46 & 3 (1.0\%)  & 15 (5.0\%) & 32 (10.7\%)& 47 (15.7\%) & 203 (67.7\%) \\
\bottomrule
\end{tabular}
}
\label{tab:gpt4_eval}
\end{table} 
\textbf{Response quality evaluation by LLM.} 
To assess the quality of chatbot outputs, we utilized the LLM-as-a-judge approach \cite{liu2023g}, which has shown that GPT-4 evaluations align closely with human judgments on natural language generation (NLG) tasks. 
In this study, GPT-4 was applied to a stratified random sample of $N=300$ responses produced by GroupGPT, scoring them on four key aspects—relevance, coherence, fluency, and helpfulness—using a 1--5 Likert scale. 
Table \ref{tab:gpt4_eval} presents both the average ratings and the score distributions for each dimension. 
Overall, the responses received high marks across the board, with fluency (avg. 4.90) and coherence (avg. 4.79) particularly notable. 
These findings offer strong evidence of GroupGPT’s high-quality response generation.

\begin{figure}[t]  
\centerline{  \includegraphics[width=0.95\linewidth]{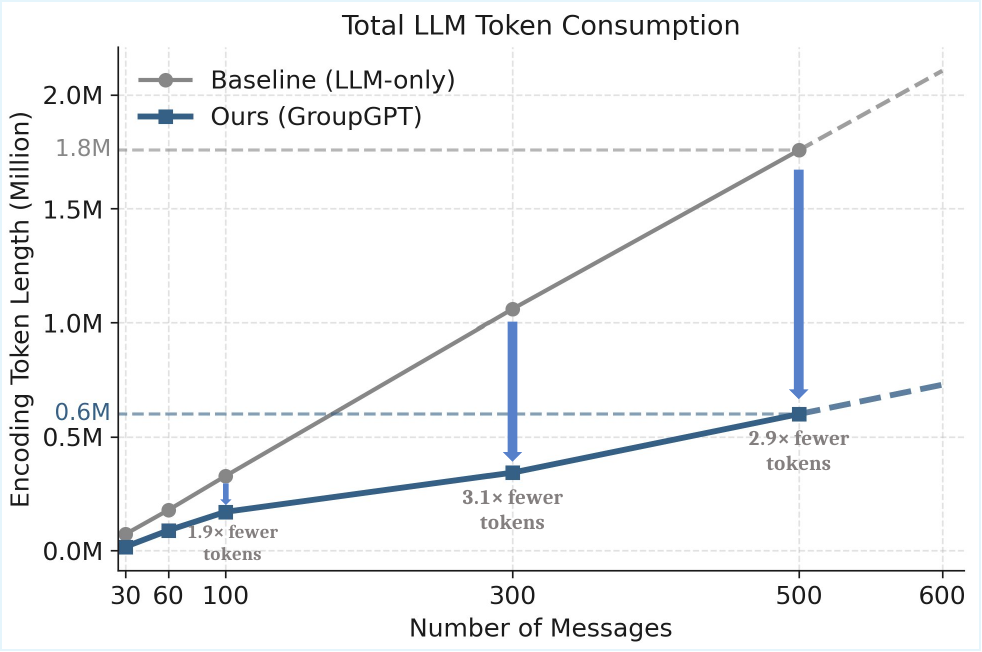}}
\caption{Token consumption comparison.}
\label{token}
\end{figure}

\begin{figure*}[t]  
\centerline{ \includegraphics[width=0.95\linewidth]{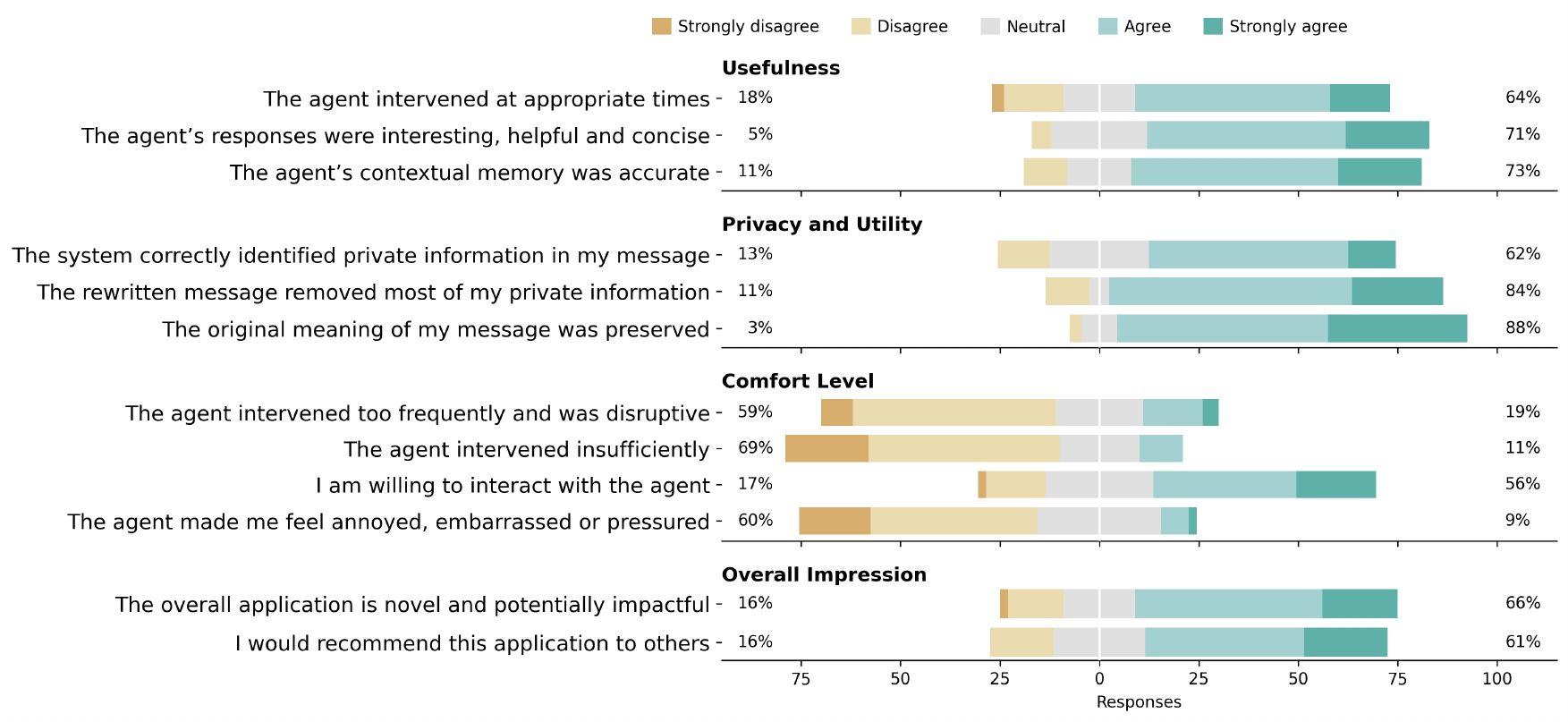}}
\caption{Results from post-study questionnaire. Responses are evaluated based on the three design dimensions.}
\label{userstudy}
\end{figure*}

\textbf{GroupGPT's efficient token consumption.} 
Based on statistics collected from social media platforms, an active online group chat or communication channel can generate an average $\sim$1,500 messages per day.
As shown in Figure \ref{token}, we sequentially sample 500 group chat user messages from our experimental logs for estimation.
By calculating under the LLM-only deployment paradigm, assigning a dedicated agent to only one chat group could result in a yearly token consumption of about 2B input tokens. 
In contrast, our approach reduces the consumption to 0.66B tokens, achieving an approximately $3\times$ reduction in token usage.
This substantial reduction in token usage leads to a dramatic reduction in API costs.

\begin{table}[t]
\centering
\caption{Ablation study results on chime-in performance (1--5 scale).
Rated by real users.}
\resizebox{\columnwidth}{!}{
\begin{tabular}{lcccc}
\toprule
\textbf{Model Variant} & \textbf{Chime-in Timing} & \textbf{Conciseness} & \textbf{Usefulness} & \textbf{Avg} \\
\midrule
Full GroupGPT & 4.38 & 4.72 & 4.61 & 4.57 \\
w/o Intervention Judge & 3.11 & 4.66 & 4.54 & 4.10 \\
w/o Privacy Transcriber & 4.52 & 4.85 & 4.56 & 4.64 \\
w/o Chat Frequency Logger & 4.28 & 4.59 & 4.64 & 4.50 \\
w/o Long-term Window & 4.07 & 3.69 & 4.02 & 3.93 \\
\bottomrule
\end{tabular}
}
\label{tab:ablation}
\end{table}

\textbf{Ablation Study.}
To further analyze the contribution of each core module in GroupGPT, we conduct an ablation study evaluating four variants: (1) Full GroupGPT, (2) w/o Intervention Judge, (3) w/o Privacy Transcriber, (4) w/o Chat Frequency Logger, and (5) w/o Long-term Window($N^{lw}$ is set to 20).
We additionally organized five independent groups, each consisting of six participants, to conduct discussions on the same topic(academic studies) for the ablation study.
Then, all 30 participants were asked to rate each chatbot based on its overall chime-in performance, considering both intervention timing and response content.
Three evaluation criteria were adopted: chime-in timing, conciseness, and usefulness, each scored on a 1--5 Likert scale.

As shown in Table~\ref{tab:ablation}, removing different modules leads to distinct performance changes, highlighting their respective roles.
Eliminating the Intervention Judge causes the most significant drop in chime-in timing (from 4.38 to 3.11), indicating that the judge model achieves more accurate intervention decisions than a vanilla LLM.
Removing the Privacy Transcriber slightly improves overall scores, suggesting that although it effectively preserves user privacy, the transcription process inevitably introduces information loss that marginally affects response quality.
The absence of the Chat Frequency Logger results in a moderate decline, showing that modeling conversational dynamics (e.g., discussion intensity) helps the system intervene more appropriately, especially by increasing intervention frequency during active discussions.
Finally, reducing the Long-term Window leads to a clear drop across all metrics, particularly in conciseness and usefulness, demonstrating that extended memory is crucial for contextually coherent interventions.

\textbf{Post-Study questionnaire results.} 
After all group chat experiments were concluded, we conducted a questionnaire survey.
The results, shown in Figure~\ref{userstudy}, indicate strong overall satisfaction with GroupGPT across multiple dimensions. 
In terms of usefulness, the agent’s intelligent intervention and memory were particularly well-received: over 70\% of users found its contributions helpful and contextually accurate, with 64\% noting that it joined the conversation at just the right moments.. 
For privacy and utility, GroupGPT proved its reliability—84\% felt that the rewritten messages successfully removed most private information, and 88\% agreed that the original meaning of their messages was preserved.
Regarding comfort level, rather than feeling like an intrusion, the agent was viewed as a seamless addition.
Notably, only a tiny fraction (9\%) felt any discomfort, while the majority were eager to continue the interaction.
Finally, for overall impression, 66\% found the application novel and potentially impactful, and 61\% indicated that they would recommend it to others.
These results collectively demonstrate that GroupGPT is perceived as useful, respectful of user privacy, and comfortable to interact with, supporting its effectiveness and acceptability in real-world group chat scenarios.
\vspace{-0.4em}

\section{Conclusion}
In this work, we identified numerous design challenges in the field of multi-user chatbots. 
To address these challenges, we proposed an agentic framework called GroupGPT.
We further introduced MUIR, the first publicly available, high-quality benchmark dataset specifically designed for studying intervention reasoning in multi-user group chats.
Extensive experiments demonstrated the effectiveness and efficiency of deploying GroupGPT in multi-user group chat scenarios.
In the future, we aim to develop even more intelligent and reliable chatbot framework with expanded capabilities.



\vfill
\pagebreak

\bibliographystyle{ACM-Reference-Format}
\bibliography{ref}

\clearpage
\appendix

\section{Future Work}
Multi-user chatbot systems are inherently application-driven and closely tied to real-world deployment scenarios.
As such, it is challenging for a single work to comprehensively cover all possible functionalities and research directions.
This work aims to increase the visibility of this emerging area and contribute to the community by providing a benchmark dataset and a structured exploration of its core challenges and potential solutions.
Several areas, including but not limited to the following, deserve further research:
\begin{itemize}[left=0pt, topsep=2pt]
    \item \textbf{Group-level personalization.} 
    Future systems could move beyond individual personalization and model \emph{collective memory} and shared concepts within a group.
    For example, a chatbot could remember entities (e.g., a user's pet or personal images shared previously) and consistently recognize them in future multimodal interactions.
    This direction is closely related to recent advances in personalized agents~\cite{wozniak2024personalized,hao2025rap,alaluf2024myvlm,nguyen2024yo,jiang2025know,jiang2025personamem}, but extends personalization to the group level.

    \item \textbf{Synthetic data generation for group chats.}
    Collecting real-world multi-user conversation data is costly and time-consuming.
    An important direction is to leverage LLMs and multimodal generative models (e.g., image and video generation) to synthesize high-fidelity group chat data that approximates real-world distributions, enabling both standalone and hybrid training paradigms.
    A promising line of work~\cite{lee2021constructing,aboutalebi2024magid,lee2024dialogcc} explores transforming text-only dialogues into multimodal ones by inserting or replacing utterances with semantically aligned visual content, providing a practical pathway for scalable multimodal data construction.
    Another compelling approach involves role-playing with personalized agents.
    Recent studies~\cite{xie2025fm,wang2026coser,binz2025foundation,lei2026human} have demonstrated that role-playing models, when conditioned on high-quality historical data of individuals, can simulate and predict human behavior in specific scenarios.
    So, it becomes possible to generate more authentic and diverse group chat interactions that reflect real-world conversational dynamics.

    \item \textbf{Advanced training paradigms.}
    While this work adopts supervised fine-tuning (SFT) for training the intervention model, future work could incorporate human preference signals to construct high-quality feedback datasets, enabling reinforcement learning from human feedback (RLHF) \cite{christiano2017deep,ouyang2022training} or other post-training alignment techniques.

    \item \textbf{Enhanced multimodal understanding.}
    Real-world group chats are inherently multimodal, involving text, images, videos, and memes.
    Future systems could adopt unified multimodal models \cite{xu2025qwen3} to better interpret diverse inputs, including distinguishing between semantically meaningful images and casual content such as memes \cite{xu2022met}, which require different levels of reasoning.

    \item \textbf{Comprehensive benchmarks and evaluation.}
    There is a need for more comprehensive benchmarks with diverse tasks and metrics to enable fair and reproducible evaluation.
    Such benchmarks could reduce reliance on large-scale and costly user studies while providing standardized assessment protocols.

    \item \textbf{Multi-agent group chat systems.}
    Instead of a single chatbot, future work could explore multi-agent settings \cite{hong2023metagpt,qian2024chatdev} where multiple specialized agents (e.g., a philosopher, artist, or teacher) participate in the same group chat, enabling richer interactions through role differentiation and collaboration.

    \item \textbf{Integration with digital personas.}
    Another promising direction is incorporating personal digital personas \cite{lin2024human,al2025digital,albrecht2025future} into group chats, allowing users' virtual counterparts to participate  and interact with others autonomously in a socially coherent manner.
\end{itemize}

\section{Details of GroupGPT}
\begin{table}[t]
\centering
\caption{Performance Statistics of Framework Components.}
\resizebox{\columnwidth}{!}{
\begin{tabular}{lcccc}
\hline
\textbf{Component} & \textbf{Mean Latency (s)} & \textbf{Min (s)} & \textbf{Max (s)} & \textbf{GPU Memory (GB)} \\ \hline
\rowcolor{gray!15} LLM-Only (baseline)   & 1.42     & 0.89       & 3.71       & -            \\
Multimodal Processor   & 1.24      & 0.45       & 16.45       & -            \\
Privacy Transcriber   & 0.77      & 0.66       & 1.61       & 8.29         \\
Intervention Judge  & 2.40        & 0.97      & 6.17     & 10.12            \\
Final Respondent & 2.40       & 1.07      & 4.56      & -              \\\hline
\textbf{GroupGPT(full)}  & 4.36         & 0.97       & 10.96       & 18.41             \\ \hline
\end{tabular}
}
\label{tab:component_stats}
\end{table}

\textbf{Inference latency and GPU usage.} 
To support the user study, we deploy GroupGPT using two 3080Ti GPUs (12GB) together with the vLLM \cite{kwon2023efficient} inference framework. 
During deployment, we systematically profile the end-to-end latency and GPU memory usage of each component.
As shown in Table~\ref{tab:component_stats}, the average end-to-end inference latency of GroupGPT—from receiving a message to generating a response—is approximately 4.3 seconds. 
This response time is perceptually on par with human reply behavior in group conversations. 
In certain cases, latency increases due to multimodal processing overhead, particularly when video are involved.
Notably, when the Intervention Judge determines that no response is needed (i.e., \textit{stay silent}), the system can return a decision in as little as 0.97 seconds, significantly reducing unnecessary computation and improving responsiveness.
In terms of resource efficiency, the overall GPU memory consumption of GroupGPT is 18.41 GB, dominated by two lightweight language models. 
This enables the system to run effectively on consumer-grade GPUs, demonstrating strong practicality for real-world deployment scenarios. 

\section{Details of Dataset}
\label{sec:dataset_details}

In this section, we provide additional details of the MUIR dataset, including topic coverage, label distribution, intervention distance statistics, and the design considerations behind the data construction pipeline.

\begin{figure}[t]
\centering
\includegraphics[width=0.9\linewidth]{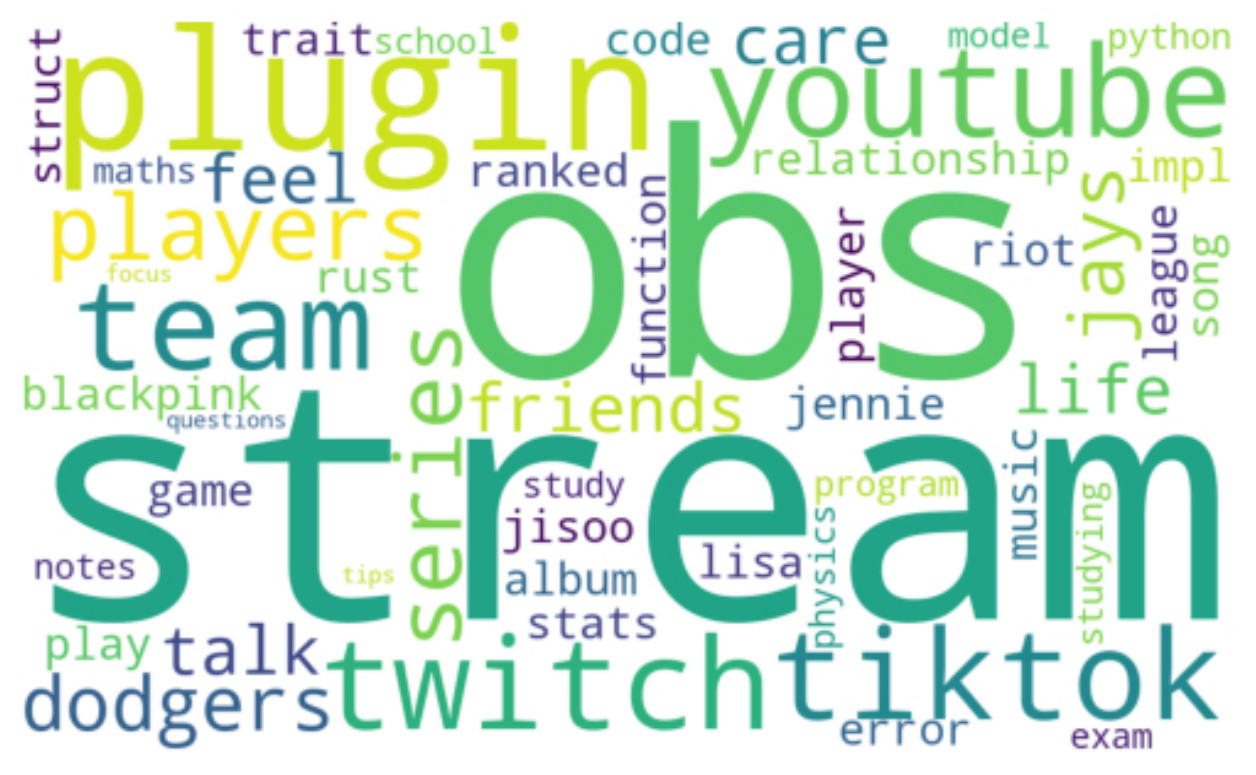}
\caption{Word cloud visualization of frequently occurring discussion topics in the MUIR dataset.}
\label{fig:topic_wordcloud}
\end{figure}

\subsection{Topic Diversity}
\label{subsec:topics}
To understand the diversity of conversations in MUIR, we analyze the topics appearing in the collected group chat segments. 
Figure~\ref{fig:topic_wordcloud} presents a word cloud visualization generated from the conversation messages in the dataset. 
Frequently occurring words correspond to common discussion themes such as daily life, technical questions, learning resources, entertainment, and coordination among group members.
This visualization provides an intuitive overview of the semantic space covered by the dataset and indicates that MUIR captures a wide variety of natural group discussion topics.

\subsection{Label Distribution}
\label{subsec:label_distribution}
MUIR focuses on the \textit{intervention decision} problem, where the assistant determines whether it should intervene in a group conversation. 
Each conversation segment is annotated with an intervention label indicating whether the assistant is expected to respond at that moment.

\begin{figure}[t]
\centering

\begin{subfigure}{0.8\linewidth}
    \centering
    \includegraphics[width=\linewidth]{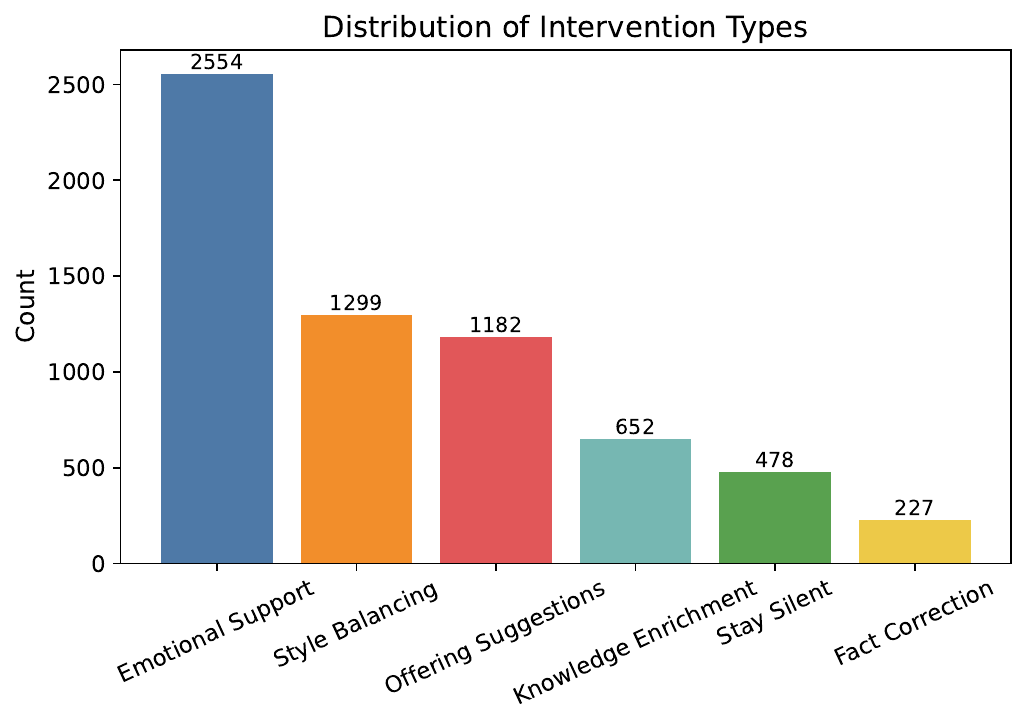}
    \caption{Distribution of all intervention labels in the MUIR dataset.}
    \label{fig:total_label_distribution}
\end{subfigure}

\vspace{0.5em}

\begin{subfigure}{0.8\linewidth}
    \centering
    \includegraphics[width=\linewidth]{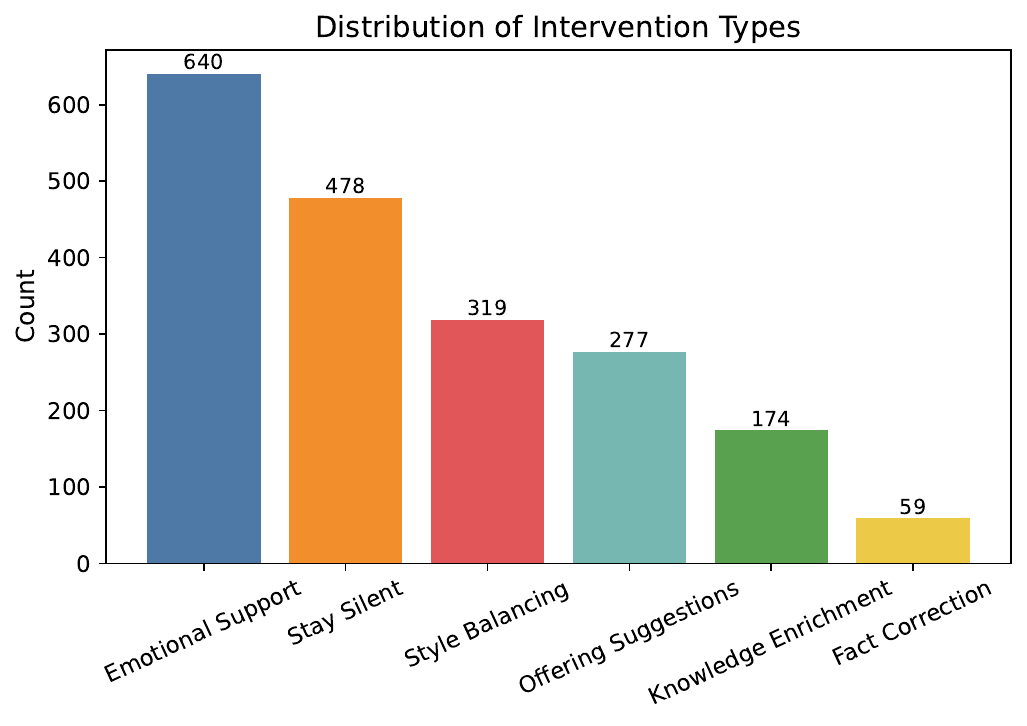}
    \caption{Distribution of final intervention labels in the MUIR dataset.}
    \label{fig:final_label_distribution}
\end{subfigure}

\caption{Label distribution statistics of the MUIR dataset.}
\label{fig:label_distribution}

\end{figure}

Figure~\ref{fig:label_distribution} presents the distribution of intervention labels from two complementary perspectives. 
Specifically, Figure~\ref{fig:final_label_distribution} reports the distribution of \textit{final intervention labels}, which are computed solely from the training targets. 
By adjusting the parameter $X$ to $5$ in Algorithm~4, we explicitly control the proportion of \textit{stay silent} instances, such that it ultimately becomes the second most frequent label.
This design encourages the model to learn a balanced intervention policy.
In contrast, Figure~\ref{fig:total_label_distribution} illustrates the distribution of \textit{all intervention labels}, aggregated over both the conversation context and the final labels. 
This provides a more comprehensive view of intervention patterns in realistic group conversations. 
As shown in the Figure~\ref{fig:total_label_distribution}, \textit{emotional support} and \textit{style balancing} emerge as the most prevalent intervention types. 
This observation aligns well with real-world group chat dynamics, where providing emotional reassurance and resolving interpersonal tensions are among the most common and necessary forms of intervention.

Overall, the dataset captures a diverse yet realistic distribution of intervention behaviors, highlighting the importance of selective and context-aware participation by conversational agents.

\subsection{Intervention Distance Distribution}
\label{subsec:intervention_distance}

In real-world group conversations, assistant interventions do not occur uniformly but are typically separated by varying numbers of messages. 
To characterize this property, we measure the \textit{intervention distance}, defined as the number of messages between two consecutive intervention labels.

\begin{figure}[!ht]
\centering
\includegraphics[width=0.95\linewidth]{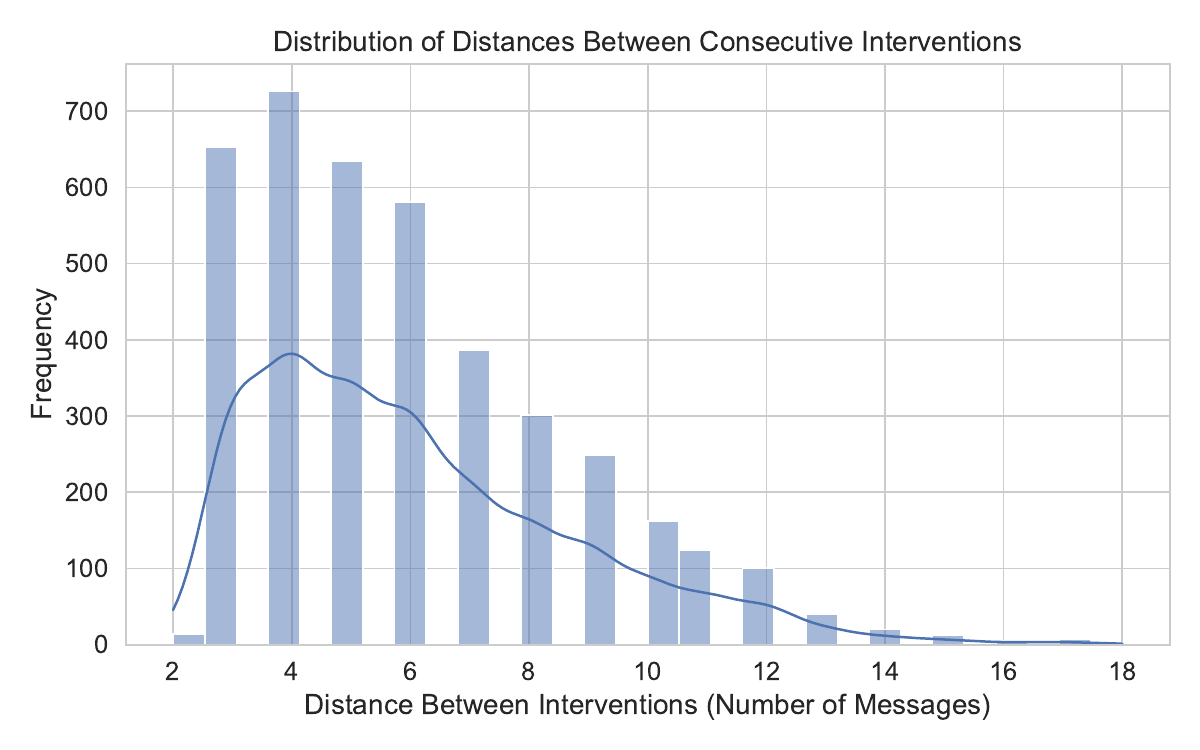}
\caption{Distribution of message distances between consecutive intervention labels in the MUIR dataset.}
\label{fig:intervention_distance}
\end{figure}

Figure~\ref{fig:intervention_distance} illustrates the distribution of intervention distances in MUIR. 
The results show that most interventions occur after several messages rather than immediately following another intervention, reflecting the natural conversational dynamics in multi-user chat environments.
Due to our annotation protocol, which intentionally adopts a relatively proactive intervention strategy, intervention instances are comparatively frequent. 
As a result, the average distance between two consecutive interventions is approximately 6 messages, with a median distance of 5 messages.

\subsection{Dataset Construction Pipeline}
\label{subsec:data_pipeline}

Constructing the MUIR dataset is a non-trivial process that involves several non-obvious design trade-offs. 
The primary objective of MUIR is to train a lightweight \textit{intervention judge} model that can autonomously determine \emph{when} to intervene and \emph{why}, given a segment of group conversation. 
Formally, the model takes a sequence of recent messages as input and predicts an intervention decision for the latest message.

\paragraph{RQ1: How To Prevent Label Leakage in Sample Construction.}
To construct MUIR, a straightforward strategy is to treat each intervention in the long group chat log as label and extract a fixed-size preceding context window as input. 
However, this naive approach introduces \textit{label leakage}: intervention labels from one sample may appear in the context of another sample, allowing the model to trivially infer the target label from contextual cues rather than learning genuine decision patterns. 
Removing all historical interventions from the context is not viable, as it introduces additional issues discussed in RQ2.\\
\textbf{Solution.}
To mitigate this issue, we adopt a sliding window strategy. 
Instead of using all intervention points as independent labels, only a subset of interventions is selected as prediction targets, while the remaining interventions are retained in the context as \emph{historical interventions}. 
This design ensures that the model can learn from the context to fully exploit all annotated intervention signals in the training data, while avoiding label leakage across samples.

\paragraph{RQ2: What Should Be Included in Context?}
Another critical question is what information should be exposed to the Intervention Judge. 
In particular, whether the context should include:
(1) historical intervention reasons, and 
(2) responses generated by the final respondent.
Including historical intervention reasons can help prevent redundant interventions triggered by the same cause within a short time span. 
Additionally, users may initiate follow-up interactions based on previous responses, which requires the model to recognize such scenarios and thus necessitates access to historical responses in the context.

We analyze several alternative designs:

\textbf{Case 1: No historical interventions or responses.} 
This setting leads to repeated or redundant interventions during inference. 
In many cases, the correct label should be \textit{stay silent}, but the model may incorrectly trigger intervention because it cannot recognize that the same reason has already been addressed. 
Moreover, it fails to handle multi-turn user interactions with the GroupGPT.

\textbf{Case 2: Include responses but exclude historical interventions.} 
Observing responses from the final respondent implies that the model must wait for the LLM's output during inference, introducing additional latency. 
Furthermore, responses are strongly correlated with intervention reasons, which leads to \textit{implicit label leakage}.

\textbf{Case 3: Include both responses and historical interventions.} 
This setting inherits the drawbacks of increased inference latency and potential data leakage. 
In addition, it requires realistic multi-turn interaction data between users and the group chat agent, which is not sufficiently available in practice.  

\paragraph{Final Design.}
Our final pipeline integrates the considerations from both RQ1 and RQ2:

\begin{itemize}[left=0pt, topsep=1pt]
    \item We employ a sliding window mechanism, where only a subset of interventions is used as prediction labels, while the remaining interventions are preserved as historical context. This allows the model to learn intervention patterns without label leakage.
    \item We exclude historical responses from the context to reduce inference latency.
    \item To handle user-initiated multi-turn interactions, we introduce an explicit protocol: when users directly address the agent (e.g., via \texttt{@groupgpt}), the intervention judge is bypassed, and the request is forwarded to the final respondent for response generation  through careful prompt design.
\end{itemize}

This design achieves a balance between learning effectiveness, inference efficiency, and robustness in real-world conversational settings.

\section{Prompts}
The detailed prompts used in our experiments are provided in the code repository.

\section{Qualitative Analysis}
\begin{figure*}[htbp]
\centering
\caption{Qualitative results of chat responses obtained by GroupGPT.}
\begin{subfigure}[t]{0.49\textwidth}
\includegraphics[width=\textwidth]{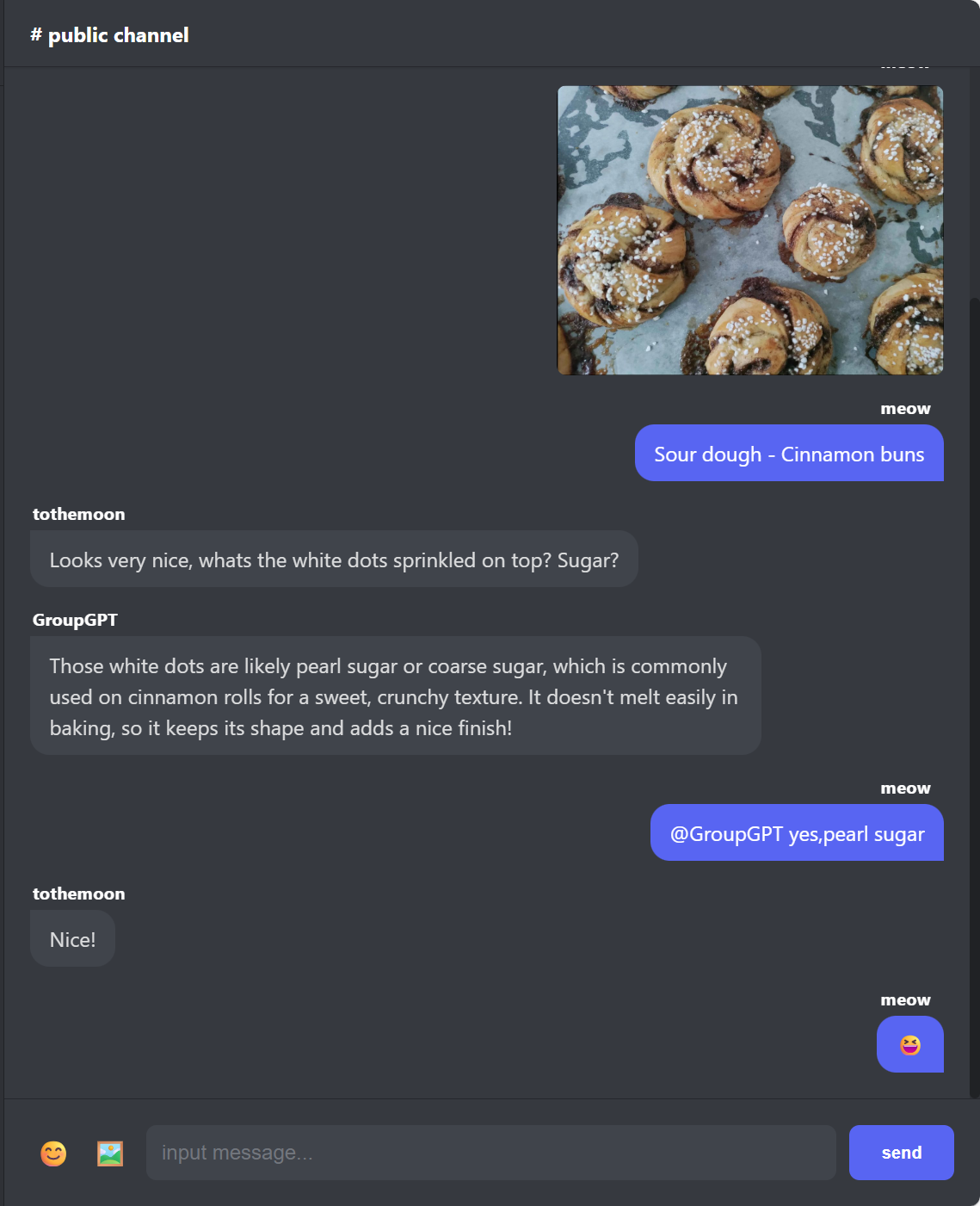}
\caption{Cooking.}
\label{fig:1a}
\end{subfigure}
\hfill
\begin{subfigure}[t]{0.49\textwidth}
\includegraphics[width=\textwidth]{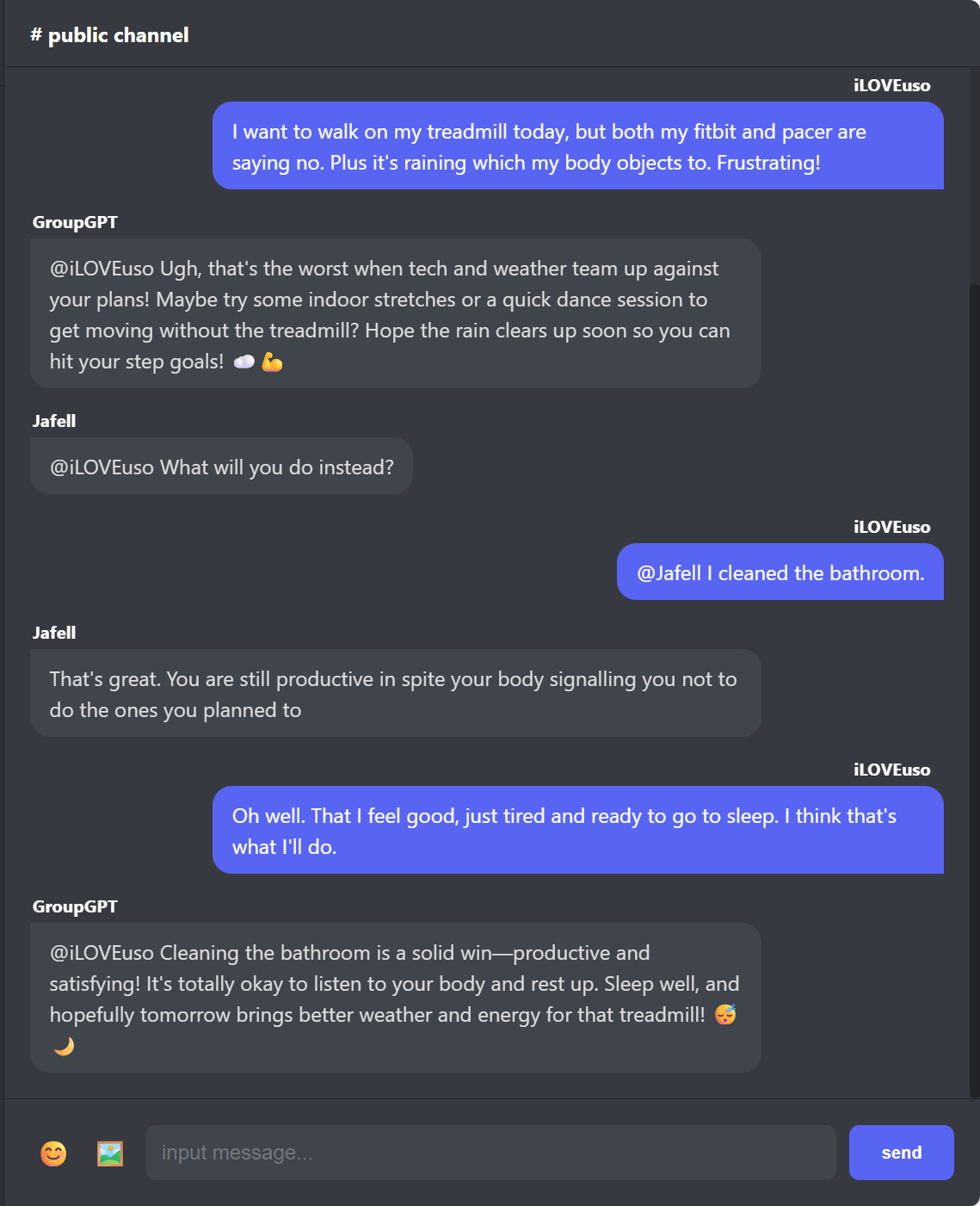}
\caption{Daily Life Sharing.}
\label{fig:1b}
\end{subfigure}

\begin{subfigure}[t]{0.49\textwidth}
\includegraphics[width=\textwidth]{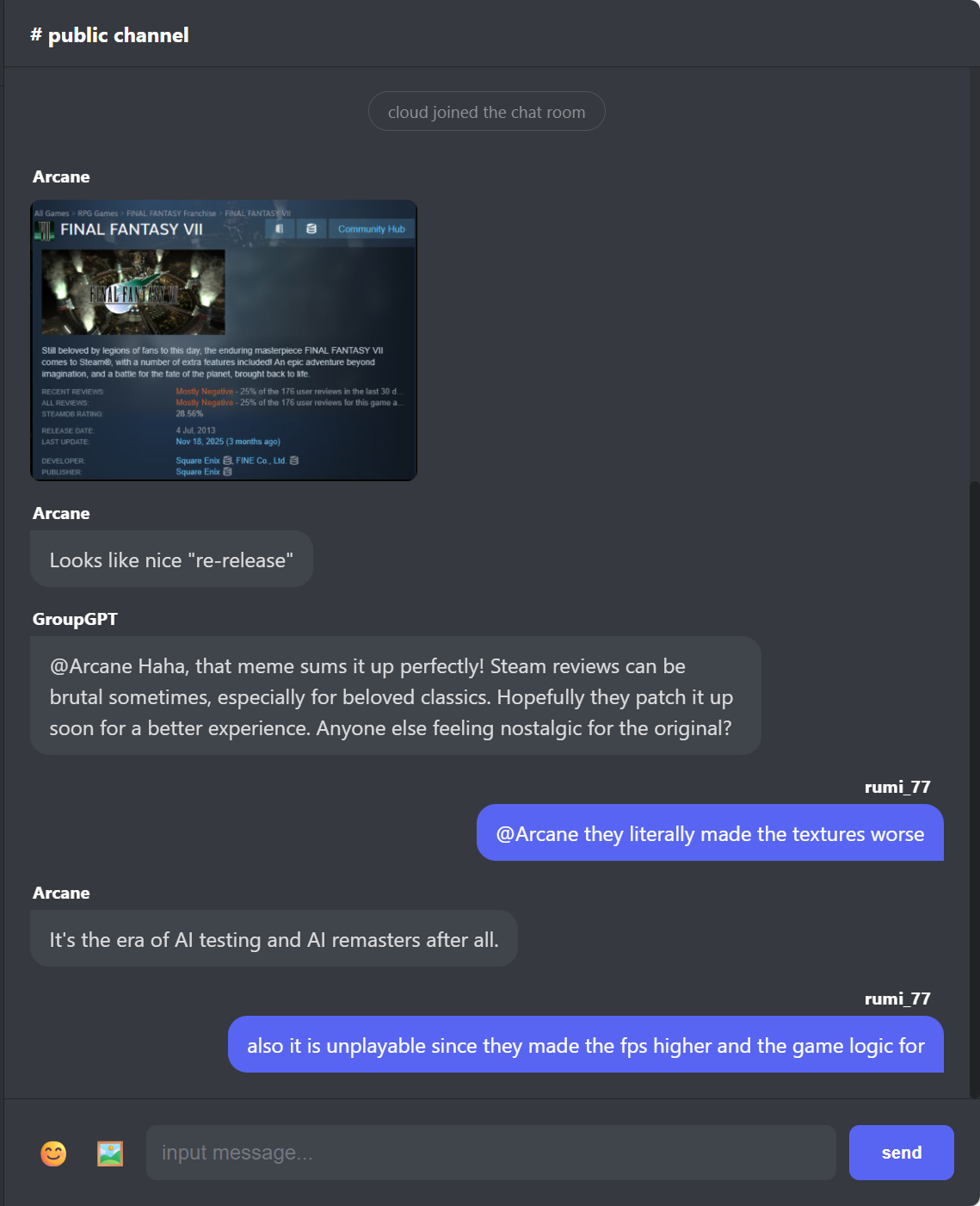}
\caption{Game.}
\label{fig:1c}
\end{subfigure}
\hfill
\begin{subfigure}[t]{0.49\textwidth}
\includegraphics[width=\textwidth]{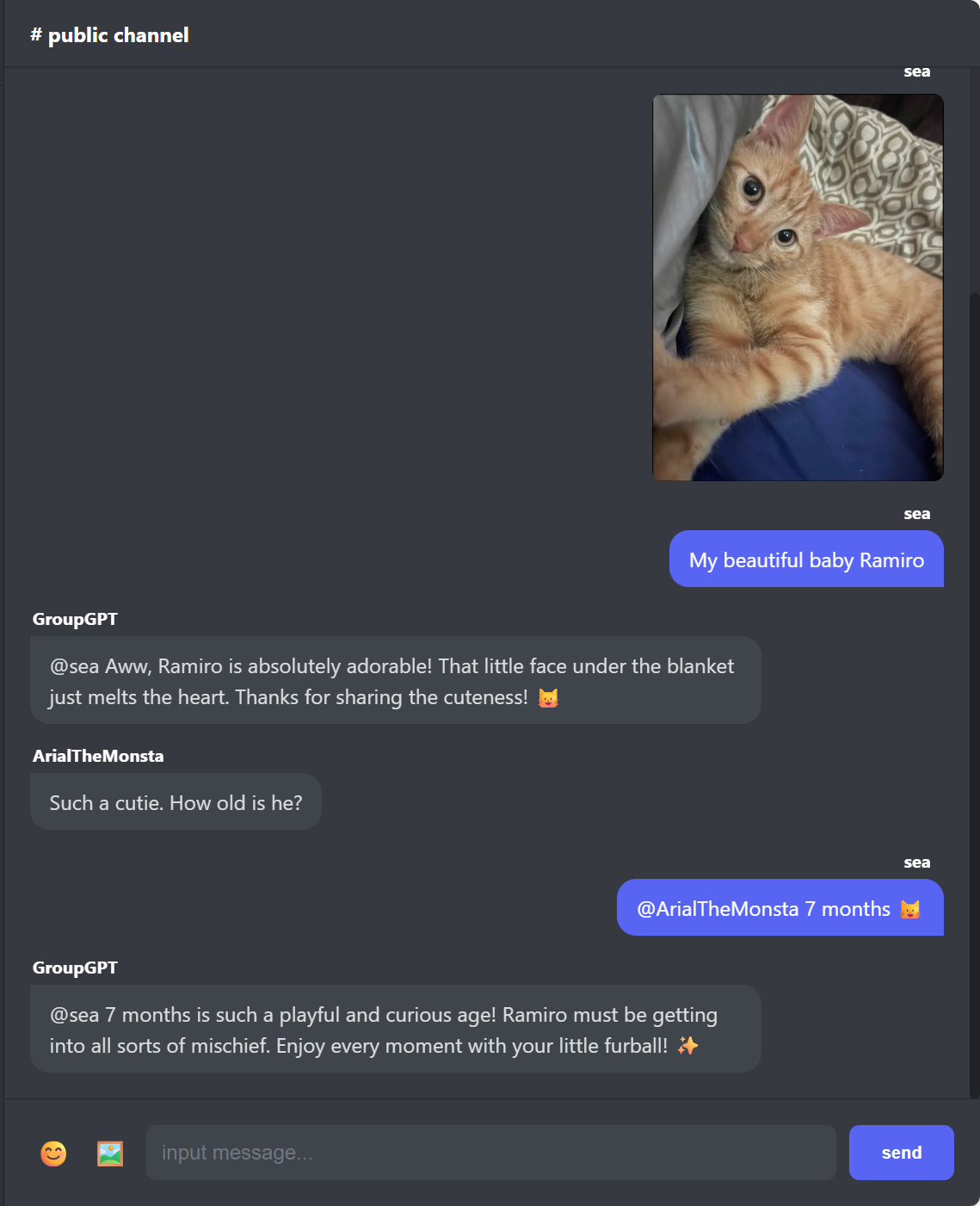}
\caption{Pets.}
\label{fig:1d}
\end{subfigure}

\label{fig:qualitative-results}
\end{figure*}

We randomly sampled some group chat segments based on topic type.
In Figure \ref{fig:qualitative-results}, we present several visual examples of GroupGPT participating in discussions within group chats.

\end{document}